\documentclass[journal]{IEEEtran}

\ifCLASSINFOpdf
\else
\fi

\usepackage{url}
\usepackage[colorlinks,linkcolor=black, anchorcolor=black, citecolor=black,urlcolor=black]{hyperref}
\usepackage{caption}
\usepackage{stmaryrd}
\usepackage{amsfonts}
\usepackage{graphicx}
\usepackage{amsmath,bm}
\usepackage{caption}
\usepackage{graphicx,times,amsmath} 
\usepackage[ruled,vlined]{algorithm2e}

\SetCommentSty{mycommfont}
\newtheorem{myDef}{Definition}
\makeatletter
\newcommand{\algorithmfootnote}[2][\footnotesize]{%
  \let\old@algocf@finish\@algocf@finish
  \def\@algocf@finish{\old@algocf@finish
    \leavevmode\rlap{\begin{minipage}{\linewidth}
    #1#2
    \end{minipage}}%
  }%
}
\makeatother

\usepackage{threeparttable}
\usepackage[numbers,sort&compress]{natbib}

\begin{document}

\title{\ \\ \LARGE\bf ModuleNet: Knowledge-inherited Neural Architecture Search}
\author{Yaran Chen, Ruiyuan Gao, Fenggang Liu and Dongbin Zhao$^{*}$~\IEEEmembership{Fellow,~IEEE}
\thanks{Y. Chen, and D. Zhao are with The State Key Laboratory of Management and Control for Complex Systems, Institute of Automation, Chinese Academy of Sciences, Beijing 100190, and also with the College of Artificial Intelligence, University of Chinese Academy of Sciences, Beijing 100049, China. (email: chenyaran2013@ia.ac.cn, lihaoran2015@ia.ac.cn, dongbin.zhao@ia.ac.cn) }
\thanks{R. Gao is with Beihang Univercity, Beijing, China. (email: gaoruiyuan@buaa.edu.cn) }
\thanks{F. Liu is with Beijing Institute of Technology, Beijing, China. (email: liufgtech@bit.edu.cn) }
\thanks{$^{*}$: D. Zhao is the corresponding author}}


\markboth{Journal of \LaTeX\ Class Files,~Vol.~14, No.~8, April~2020}%
{Shell \MakeLowercase{\textit{et al.}}: Bare Demo of IEEEtran.cls for IEEE Transactions on Magnetics Journals}


\maketitle
\IEEEpeerreviewmaketitle


\begin{abstract}
    Although Neural Architecture Search (NAS) can bring improvement to deep models, they always neglect precious knowledge of existing models.
    The computation and time costing property in NAS also means that we should not start from scratch to search, but make every attempt to reuse the existing knowledge.
    In this paper, we discuss what kind of knowledge in a model can and should be used for new architecture design.
    Then, we propose a new NAS algorithm, namely ModuleNet, which can fully inherit knowledge from existing convolutional neural networks.
    To make full use of existing models, we decompose existing models into different \textit{module}s which also keep their weights, consisting of a knowledge base.
    Then we sample and search for new architecture according to the knowledge base.
    Unlike previous search algorithms, and benefiting from inherited knowledge, our method is able to directly search for architectures in the macro space by NSGA-II algorithm without tuning parameters in these \textit{module}s.
    Experiments show that our strategy can efficiently evaluate the performance of new architecture even without tuning weights in convolutional layers.
    With the help of knowledge we inherited, our search results can always achieve better performance on various datasets (CIFAR10, CIFAR100) over original architectures.
 \end{abstract}
 

\section{Introduction}

Convolutional Neural Networks (CNN) have been successfully applied to various computer vision tasks, such as image classification~\cite{AlexNet,ResNet,VGG,D.Zhao} and object detection~\cite{FPN,FRCNN}.
All these impressive results thank to human experts' discovery of finer architectures and design principles, but cost too much effort.
Nowadays, manual design can hardly satisfy the increasing needs of various applications.

\begin{figure}[t]
   \centerline{
      \includegraphics[width=3.28in]{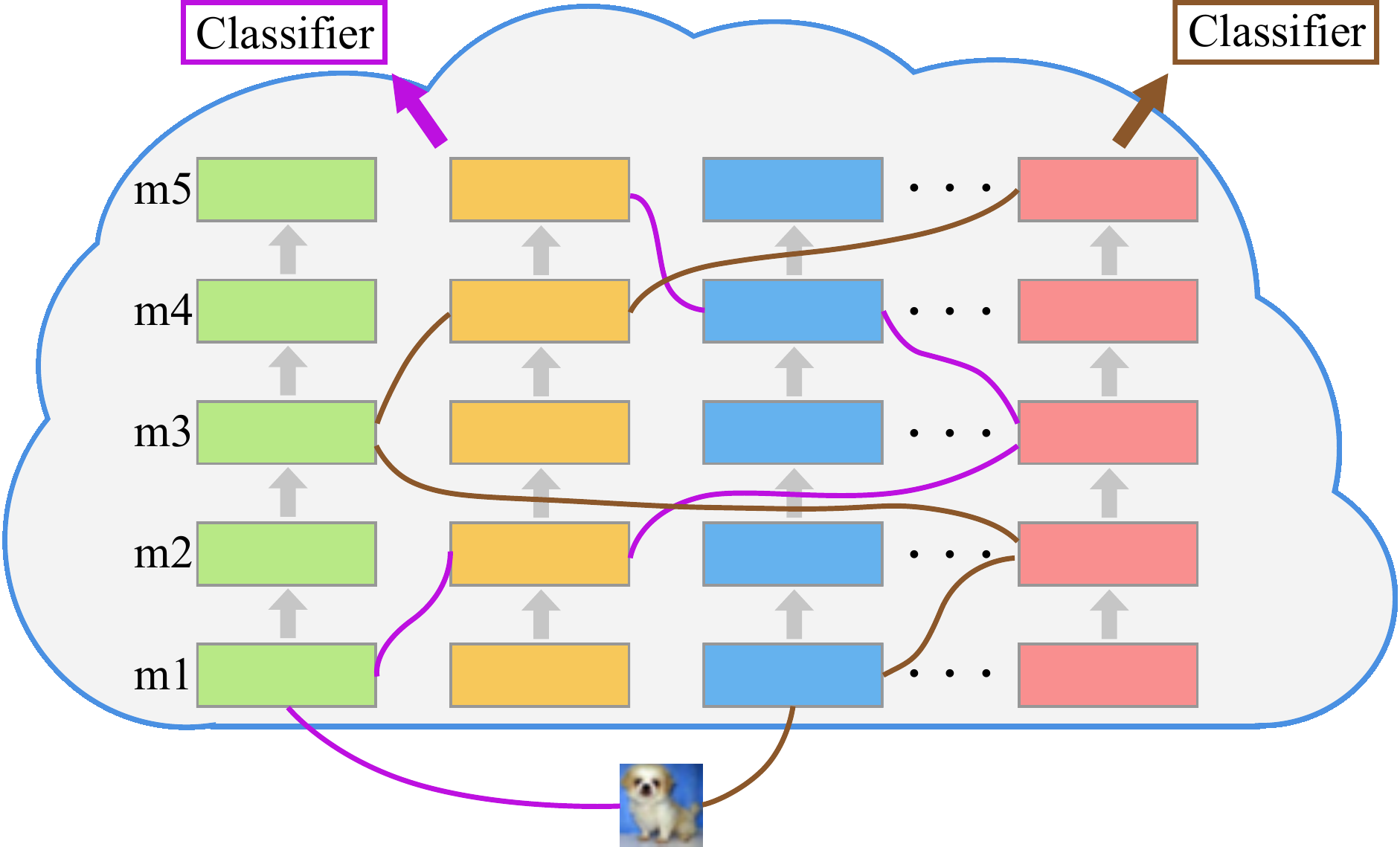}
   }
   \caption{An example of architecture generation of ModuleNet. We decompose some existing architectures (shown with the same color) into 5 cells  and keep their parameters to form \textit{module}s (``m'' in the figure).
   Arrows in grey show original inference paths.
   Purple lines and brown lines separately show two possible architectures.}
   \label{fig:instance}
\end{figure}

Auto Machine Learning (AutoML) provides an efficient paradigm to automate model design.
Especially, Neural Architecture Search (NAS) algorithms achieve the optimization of architecture design on a dataset of interest.
Current NAS algorithms are generally three folded.
First, search for convolutional layers (or ``cells'') considering a given task.
Second, repeatedly stack a searched cell for several times to integrate a deep enough architecture.
Finally, fine-tune this architecture on the target dataset. 
Limited by its computation demanding nature, NAS can find new architectures that exceed the best performance of manual design~\cite{ENAS,DARTS,Li} only with high-performance hardware.
Therefore, to free NAS from resources consuming on algorithm level is of significance.

However, current NAS procedure only utilizes very little knowledge from experience by manual design, such as repeating cells of the same motif which consists of a combination of several operators~\cite{NASNet,Ding}.
However, given a trained CNN, there are at least two aspects of knowledge which can be used extensively, thereby, reducing the search cost. 

\textbf{First, architecture}: 
Every improvement of CNN contains precious knowledge, which reflects scientists' comprehension of CNN and inspirations from that.
Starting from AlexNet~\cite{AlexNet}, CNNs have made great progress in computer vision tasks.
The success of VGG~\cite{VGG} confirms the significance of depth in visual representations.
Kaiming He's introduction of shortcut connection in ResNet~\cite{ResNet} saves CNNs from the degradation problem when going deeper.
GoogLeNet~\cite{GoogLeNet}, UNet~\cite{UNet} and FPN~\cite{FPN} separately show great importance of features in multi-scale and multi-resolution.
All these expert knowledge have great potential for rediscovery and reorganization.
However, current AutoML methods constrain themselves in searching from scratch, turning a blind eye to this knowledge.

\textbf{Second, trained parameters}: 
By training on a given dataset, CNN can learn and distill knowledge contained implicitly inside massive data.
For one thing, transfer learning through weight sharing is widely accepted in various computer vision tasks, such as backbones in object detection models~\cite{FRCNN,Y.Chen}.
Besides, weight-sharing is used as a basic method in NAS after~\cite{ENAS}.
Therefore, trained weights have great \textit{transferability}.
For the other, within a specific architecture, different trained parameters can extract features from different aspects.
Since different features are clearly helpful to separability among inputs, parameters are important for \textit{module}s to possess \textit{diversity}.
\textit{transferability} makes trained parameters usable for reorganization.
\textit{diversity} can introduce more knowledge into our consideration when searching.
Therefore, trained parameters in \textit{module}s are helpful for NAS.

From Evolution Algorithm (EA)~\cite{AmoebaNet,NSGA-Net} to Reinforcement Learning (RL)~\cite{NASNet,ENAS,Ding,Li} and gradient-based methods~\cite{DARTS}, scientists overrate optimization in the scenario of starting from the very stage to search, but overlook precious knowledge in existing architecture and trained parameters.
Actually, we should make progress by ``standing on the shoulders of giants''.

Therefore, we proposed a new NAS algorithm, namely ModuleNet, to solve the problems above. 
An example seen in Fig.~\ref{fig:instance}, we build a knowledge base for existing architectures with their trained parameters.
By searching over different \textit{module}s for the whole architecture, ModuleNet can inherit all knowledge from the knowledge base. 
Specifically, we first acquire the knowledge base by decompose various architectures with their trained weights into different \textit{module}s to keep their integrity.
Then, we iteratively search for some best architectures according to the knowledge base using NSGA-II algorithm~\cite{NSGA-II}, without tuning parameters in convolutional layers.
In each iteration, new architectures will be generated by reorganizing \textit{module}s, which keep their weights as the origin to inherit from the knowledge base.
In this way, we can make full use of the existing architecture and trained parameters, rediscover and reorganize them for better results.
In our experiments, the effectiveness of ModuleNet is varified on various vision datasets and show improvements over the original architectures it inherits.

To sum up, our contributions in this paper are mainly as follows:
\begin{itemize}
   \item Analyses into existing models and useful ideas to reuse them for NAS.
   \item A new NAS algorithm to search network architectures from macro aspect, which fully inherits existing knowledge and generates new ones.
   \item An easily extended NAS paradigm for multi-objective search using NSGA-II.
\end{itemize}


\section{Related Work}

\subsection{CNN Architecture Design}
From the very step of CNN architecture design, scientists use trial-and-error to discover better architecture for target tasks.
In this stage, though laborious, various successful contributions are made, such as VGG~\cite{VGG}, ResNet~\cite{ResNet}, GoogLeNet~\cite{GoogLeNet}.
Besides, new operators and principles are also introduced for different targets.
For example, batch normalization~\cite{BN} helps us to solve the internal covariate shift.
Dense connection~\cite{DenseNet} extends thinking in skip connection to every layer in macro space.
Underlaying mechanism is discussed further in ~\cite{Pre-ResNet}, through which pre-activation architecture is discovered.
To another end, depthwise separable convolution is extendedly used to shrink the barrier between accuracy and latency~\cite{mobilenets}.
Due to the incomplete comprehension of the underlying mechanisms of CNN, however, these works can only pay attention to few aspects of CNN design.
Actually, both their inspirations of architecture design and trained parameters contain very meaningful knowledge. 
We should consider from a more general view of every part of them.

For another, with the boom of computing power by accelerating hardware, AutoML has become usable to search for promising architecture automatically.
With the help of parameter sharing and performance prediction, ENAS~\cite{ENAS} sets a good example in this area.
Although based only on one design principle from manual experience -- similar cell repeating, AutoML has been broadly developed. 
DARTS~\cite{DARTS} relaxed the search space to be continuous and make architecture generation optimizable using gradient.
Besides, various algorithms are proposed to better search for optimal architecture.
Progressive shrinking makes it possible to train a once-for-all weight before searching~\cite{once-for-all}.
Prediction with Experts Advice (PEA) theory is introduced in ~\cite{xnas} to optimize regret for better architecture search.
However, none of these works can efficiently consider previous experts' effort in architecture design, causing a huge waste.

\subsection{Evolution Algorithm for NAS}
RL~\cite{NASNet,ENAS}, EA~\cite{NSGA-Net, AmoebaNet} and gradient-based algorithm~\cite{DARTS} are always used for NAS.
Among them, EA has been used for neural network design for some time.
NeuroEvolution of Augmenting Topologies (NEAT) algorithm~\cite{NSGA-Net42}, which can only search well for small networks, could be considered as the first.
From then, various works tend to extend the usage of evolution algorithm in NAS, such as CoDeepNEAT~\cite{NSGA-Net32} or AmoebaNet~\cite{AmoebaNet}.

Conceptually, search back end of the proposed ModuleNet is inspired by NSGA-Net~\cite{NSGA-Net}, which also uses Nondominated Sorting Genetic Algorithm II (NSGA-II)~\cite{NSGA-II} for searching.
NSGA-II is an evolutionary multi-objective optimization algorithm.
By extending NSGA~\cite{6791727}, NSGA-II solves the problem of nonelitism approach and lower its computational complexity, making it suitable for NAS.

\section{Method}
\subsection{Overview For ModuleNet}
An overview of our method can be seen in Algorithm~\ref{alg:Overview}.
In general, we first decompose some existing architectures to different cells.
Then we extract their weights to form \textit{module}s, and add them to the knowledge base.
Finally, we make use of NSGA-II algorithm as back end for searching.
In the following parts, we will focus on four important details in our proposed method.  

\begin{algorithm}[htbp]
   \caption{Search Algorithm for ModuleNet}
   \label{alg:Overview}
   \LinesNumbered 
   \algorithmfootnote{$^\dag$ Evolution procedures show respect to NSGA-II.\\
   $^\ddag$ Our proposed mothods seen in the following sections.}
   \KwIn{$n$ architectures $arch_{1}...arch_{n}$, cell number $c$, evolution generation $gen$, population size $p\_size$}
   \KwOut{population $pop$}
   decompose each $arch_{i}$ into $c$ cells, $arch_{i}\text{-}cell_{j}$ stands for $j^{\text{th}}$ cell in $arch_{i}$\;
   \tcc{Initialize knowledge base}
   \For{$j$ \textup{from} $1$ \textup{to} $c$}{
      \For{$i$ \textup{from} $1$ \textup{to} $n$}{
         $module^{i}_{j}=f_{m}(arch_{i}\text{-}cell_{j})^\ddag$\;
         $knowledge\_base$[$j$][$i$] = $module^{i}_{j}$\;
      }
   }
   \tcc{Initialize population}
   \For{$i$ \textup{from} $1$ \textup{to} $p\_size$}{
      \For{$j$ \textup{from} $1$ \textup{to} $c$}{
         $pop$[$i$][$j$] = $module^{*}_{j}$ sampled from $knowledge\_base$[$j$][:]\;
      }
   }
   evaluate$^\ddag$ individuals in $pop$\;
   \tcc{Do evolution search}
   \For{$g$ \textup{from} $2$ \textup{to} $gen$}{
      $new\_individual$ = mate$^\dag$ and generate$^\dag$ according to $pop$ and encoding method$^\ddag$\;
      assemble new architecture with connections$^\ddag$\;
      evaluate$^\ddag$ individuals in $new\_individual$\;
      compare$^\dag$ over $(pop+new\_individual)$ and sort$^\dag$\;
      $pop$ = select$^\dag$ from sort results\;
   }
\end{algorithm}

In Sec.~\ref{Knowledge Base}, we will illustrate how we decompose an existing architecture to make it suitable for reassembling, and compatible with other \textit{module}s in the new architectures.
We fix parameters from our knowledge base of each \textit{module} to both save much computing cost for gradient backward and effectively inherit knowledge from existing \textit{module}s.
In Sec.~\ref{Encoding}, our encoding method for different \textit{module}s, together with the definition of search space will be illustrated.
These two parts can be considered as preprocessing for search.

After that, Sec.~\ref{Module Connection} and Sec.~\ref{Performance Evaluation} serve as key points in our method.
To relieve the pressure of parameter-tuning when searching, we design new operators as connection, namely Channel Pool and Channel DePool (ChP and ChDP).
Together with fix parameter, we use these operators to completely eliminate trainable parameters before linear layers when searching.
And in Sec.~\ref{Performance Evaluation},  we will introduce a new function to better evaluate performance with restriction from fixed parameters. Experiments show fine correlative relationship between our function and test error obtained with all parameters trainable.

To be acknowledged, we only use NSGA-II with one objective in the paper.
However, taking advantage of basic design target of NSGA-II, the proposed ModuleNet can be easily extended for multi-objective search.


\subsection{Knowledge Base}\label{Knowledge Base}
\begin{figure}[htb]
   \vspace{-0.1cm}
   \centerline{
      \includegraphics[width=3.25in]{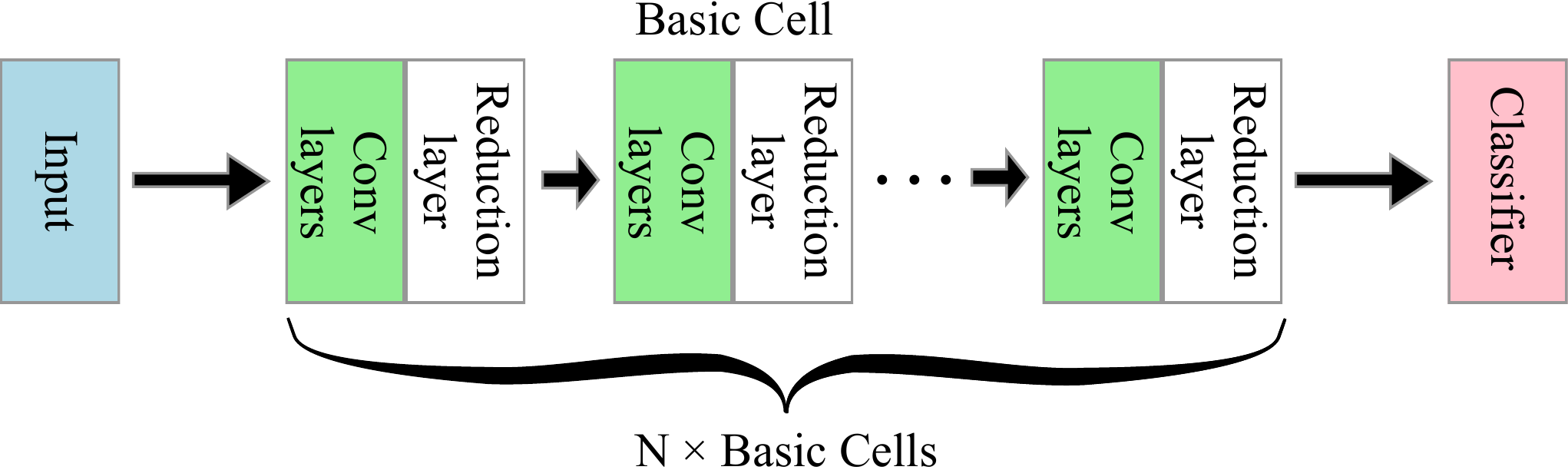}
   }
   \caption{A universal view for most CNN architectures. We combine convolutional layers and their following reduction layer as a basic cell.}
   \label{fig:uni-view}
   \vspace{-0.1cm}
\end{figure}

Existing CNN architectures, no matter discovered by experts or AutoML, are all precious knowledge that should be extendedly used.
We first decompose some existing architectures into uniform cells, and then build a knowledge base to hold.
As shown in Fig.~\ref{fig:uni-view}, inspired by \cite{ENAS}, we consider a CNN architecture as a stack of convolutional layers and reduction layers between input and classifier (always has softmax and linear layers).
Considering the continuity of the layers, we combine convolutional layers and their following reduction layer as a basic cell.

Not merely architectures of CNNs, we also consider weights in cells to avoid the burden of retraining.
Keeping weights can also help us better inherit knowledge not only from architectures, but also from training procedure.
By extracting the weight form the whole architecture, we can finally get different \textit{module}s for search, referred to as
$$
module^{i}_{j}=f_{m}(arch_{i}\text{-}cell_{j})
$$
for $j^{\text{th}}$ cell in $arch_{i}$. To make it clear, we have:
\begin{myDef}\label{def:module}
   Module: A cell decomposed from an existing CNN architecture, and keeping its trained weights in the original architecture.
\end{myDef}

Pay attention to that, in our method, we consider CNNs as multi-layer filters, and each layer can process information from different semantic aspects.
For example, layers that in a relatively shallow stage of CNNs may process information at a local level; however, deeper layers, which has a larger receptive field, fit for processing information with a global view or at a high semantic level.
Therefore, we have to keep some settings unchanged when reassembling for new architectures to make weights in \textit{module}s usable.
First, we keep the \textit{module}'s position of the order in new architecture as origin.
Second, we keep the resolution of input unchanged by adjusting the reduction in the preceding block. 
Only in this way can we make each \textit{module} take effect on its original semantic level.


\subsection{Encoding}\label{Encoding}
\begin{figure}[htb]
   \vspace{-0.1cm}
   \centerline{
      \includegraphics[width=0.8\linewidth]{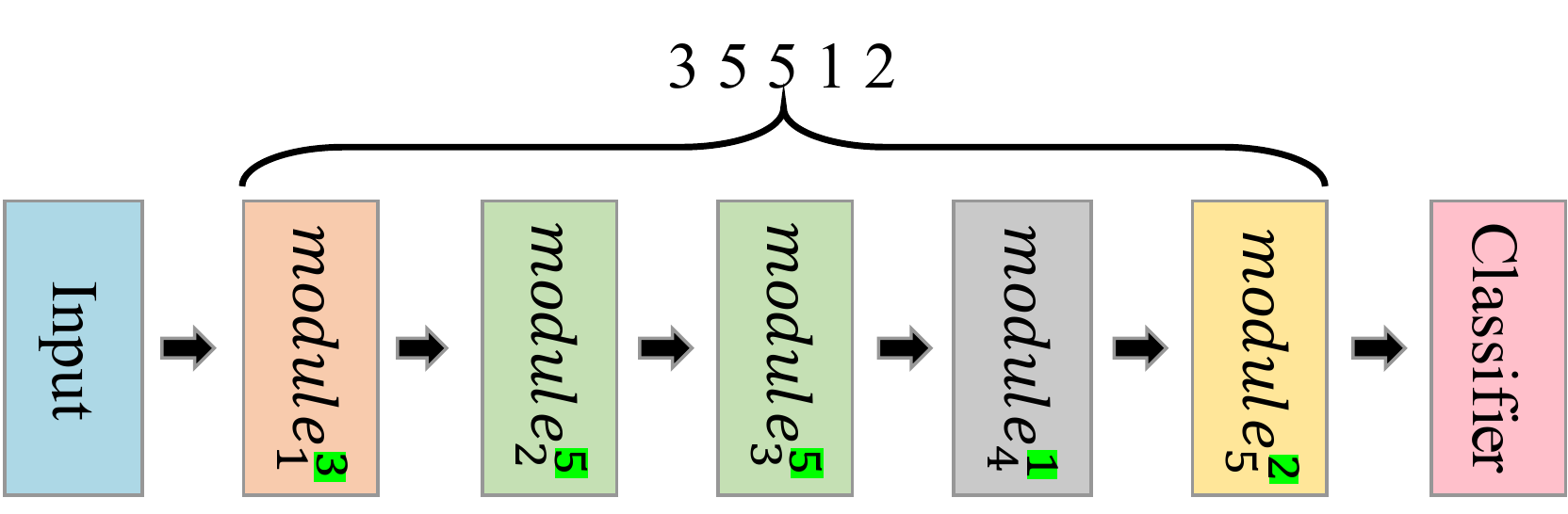}
   }
   \caption{An example of decoding string to architecture. Cells from the same architectures are in the same color.}
   \label{fig:example}
   \vspace{-0.1cm}
\end{figure}

Considering we have $n$ architectures in total, and a decomposition of $c$ cells for each architecture.
We use $module^{i}_{j}$, $\{i,j\in{N^{+}}|i\le{n},j\le{c}\}$ as representations.
By assigning different architecture $arch_{i}$ to an integer $i$, each string of integers in $\{i_{1}i_{2}...i_{c}|i\in{N^{+}},i\le{n}\}$ can be decoded as an architecture.
Specifically, $i_{j}$ represents $arch_{i_{j}}\text{-}cell_{j}$.
An example with $c=5, n\ge{5}$ is shown in Fig.~\ref{fig:example}.
Besides, we can obtain the size of search space ($\Omega$) through
$$
\setlength{\abovedisplayskip}{0.2cm}
\setlength{\belowdisplayskip}{0.2cm}
|\Omega|=n^{c}\text{,}
$$
which is much smaller than those in previous works~\cite{ENAS,DARTS,NSGA-Net} when searching in macro space, but, benefiting from existing fine design, powerful enough to make progress for tasks.

\subsection{Module Connection}\label{Module Connection}
Since we are using \textit{module}s from different architecture, which are separately designed, neighboring \textit{module}s in a new architecture may have different channels.
Following Def.\ref{def:module}, we need not and should not change weights in \textit{module}s.
However, if we use trainable parameters in connection, we still need to backpropagate loss through gradient to the front stage.
This can cause a huge computing cost, and may cause unstable due to weight-fixing in each \textit{module}.
To solve the problems above, in this section, we will introduce two new operators as connections, namely ChP and ChDP.
These two operators do not contain trainable parameters, thus solving those problems elegantly.

\textbf{Channel Pool} performs a standard 1D average pooling on the channel dimension.
It is used to decrease the number of channels.
Given $input$ with size $(N, C, H, W)$ (dimensions with the meaning of \textit{Number of batch, Channel, Height, Width}), and expected $out_{P}$ with size $(N, C_{out_{P}}, H, W)$, $C_{out_{P}} < C$, and assuming that $C_{out_{P}}|C$, we have
\begin{equation}
   out_{P}=\operatorname{ChP}(input;\{\bm{k_P}\})\text{,}
\end{equation}
where $\bm{k_P}$ is the kernel size, $\bm{k_P}=\frac{C}{C_{out_{P}}}$.
The calculation for $l^{th}$ channel dimension of $out_{P}$, or $out_{P}(*,l,*,*)$, is
$$
\begin{aligned}
&out_{P}(N_{i_{1}},l,H_{i_{2}},W_{i_{3}})\\
&=\frac{1}{\bm{k}}\sum^{\bm{k}-1}_{m=0}{input(N_{i_{1}},\bm{k}\times{l}+m,H_{i_{2}},W_{i_{3}})}\text{,}
\end{aligned}
$$
for $l\in\{l\in{N^{+}| 0\le l < \frac{C}{\bm{k_P}}}\}$.

\textbf{Channel DePool} performs a duplication and connection on the channel dimension.
It is used to increase the number of channels.
Considering $input$ with size $(N,C,H,W)$, and expected $out_{DP}$ with size $(N, C_{out_{DP}}, H, W)$, $C_{out_{DP}} > C$, and assuming that $C|C_{out_{DP}}$, we have
\begin{equation}
   out_{DP}=\operatorname{ChDP}(input;\{\bm{k_{DP}}\})\text{,}
\end{equation}
where $\{\bm{k_{DP}}\}$ is duplication times, $\bm{k_{DP}}=\frac{C_{out_{DP}}}{C}$.
The calculation for $l^{th}$ channel dimension of $out_{DP}$, or $out_{DP}(*,l,*,*)$, is
$$
\begin{aligned}
&out_{DP}(N_{i_{1}},l,H_{i_{2}},W_{i_{3}})\\
&=input(N_{i_{1}},l\text{ mod }C,H_{i_{2}},W_{i_{3}})\text{,}
\end{aligned}
$$
for $l\in\{l\in{N^{+}|0\le l < {C}\times{\bm{k_{DP}}}}\}$.

However, the actual situation may break our assumptions in definitions easily.
Therefore, we should consider more complex situations. 

For neighboring \textit{module}s that need a connection from $C$ channels to $C_{out}$ channels, if $C_{out}|C$ or $C|C_{out}$, we just use a ChP or ChDP to make connections.
Otherwise, we will extend these two operators as follows. In this part, since we only focus on the channel dimension of $input$, we use $C^{In}$ as representation. $C^{In}$ can be considered as a 1D array.

1. $C>C_{out}$\nopagebreak[4]

\begin{figure}[h]
   \centerline{
      \includegraphics[width=2.3in]{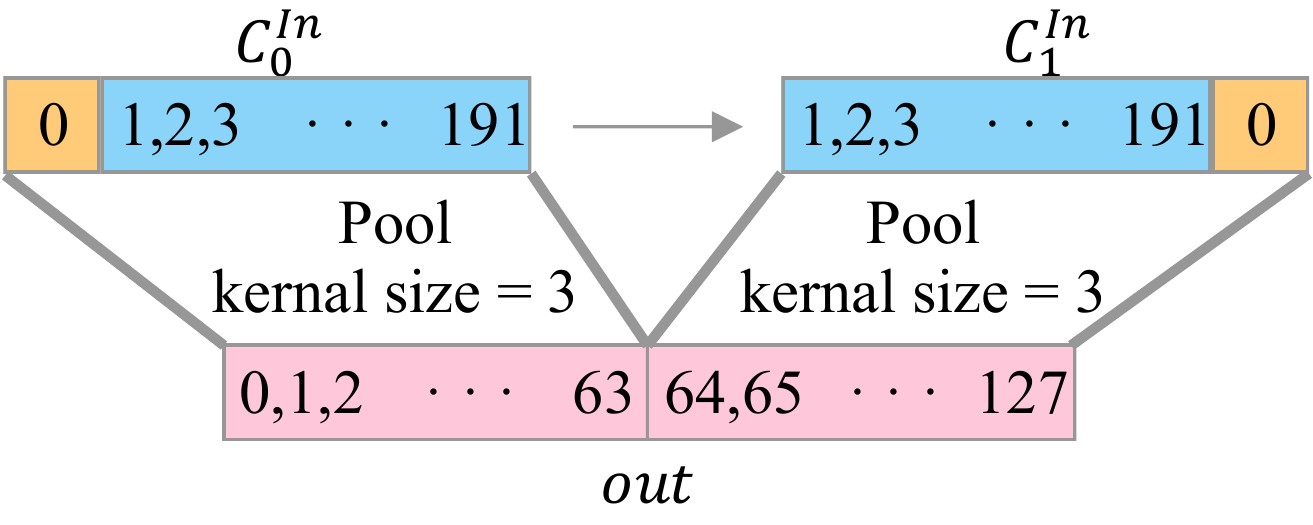}
   }
   \caption{An example for using ChP with $C^{In}=192,C_{out}=128$.}
   \label{fig:extChP}
\end{figure}

We first find the \textit{Greatest Common Divisor (GCD)} of $C$ and $C_{out}$ as $\eta$:
$$
\eta=\operatorname{gcd}(C, C_{out})\text{.}
$$
Then, we use a ChP as
\begin{equation}
   \left\{
   \begin{array}{ll}
      out&= \operatorname{Concat}\limits_{i=0}\limits^{C_{out}/\eta-1}\Big[\operatorname{ChP}(C^{In}_{i};\{\bm{k_P^{'}}\})\Big]\\
      C^{In}_{i}&=\operatorname{Concat}(C^{In}[i\text{:end}],C^{In}[\text{begin:}i])
   \end{array}
   \right.\text{,}
\end{equation}
where $\bm{k_P^{'}}=\frac{C}{\eta}$ and $[:]$ donates slice operation.
Note that $C^{In}[0\text{:end}]$ contains all elements in $C^{In}$, and $C^{In}[\text{begin:}0]$ is empty.
An example of this part can be seen in Fig.~\ref{fig:extChP}.

2. $C<C_{out}$\nopagebreak[4]

We use a slice of output from ChDP as
\begin{equation*}
   out = \operatorname{ChDP}(C^{In};\{\bm{k_{DP}^{'}}\})[\text{begin:}C_{out}]
\end{equation*}
where $\bm{k_{DP}^{'}}=\left\lceil \frac{C_{out}}{C} \right\rceil$.

Finally, through ChP and ChDP, we can make \textit{module} connections between neighbors without parameters.
In this way, the gradient can be avoided from being backpropagated deeply and much computing cost can be saved.

Experiments also show that using non-parameter connections has non-negative, even positive sometimes, effect on performance evaluation when searching.

\subsection{Performance Evaluation}\label{Performance Evaluation}
Admittedly, although fixing parameters in \textit{module}s can largely reduce both computing cost in a single iterator and total epoch needed to convergence, accuracy on the validation set ($acc_{val}$) may not fully represent the real performance for an architecture.
Since the parameters in use are extracted from some pretrained models, and these models, or architectures, are still reachable through our search algorithm, $acc_{val}$ on these models can be much higher than others.
Besides, for those architectures that are very similar to those pretrained models (only have a few \textit{module}s changed), original parameters may fit them better comparing to others with more different \textit{module}s.
To avoid this problem of unfair accuracy, we propose a new metric by taking loss changing rate ($l_{rate}$), error rate ($err_{val}$) and architecture similarity ($sim$) into consideration to better evaluate the real performance when searching, as defined by
\begin{equation}\label{equ:score}
   score=err_{val} - \alpha\cdot l_{rate} + \beta\cdot sim\text{,}
\end{equation}
where $\alpha$, $\beta$ are parameters to balance different items, and determined through experiments.
$err_{val}$ is the evaluation error on the validation set, which can be obtained by
\begin{equation}
   err_{val}=1-acc_{val}\text{.}
\end{equation}

With a basic consideration that fixing parameters may lead to a decrease in architecture's ability for generalization, but not convergence, changing rate of loss can be suitable to evaluate convergence ability of an architecture.
By defining $l_{rate}$ as the loss changing rate when training on the training set for $n$ epochs, it can be obtained by
\begin{equation}
   l_{rate}=\frac{loss_{epoch=1}-loss_{epoch=n}}{loss_{epoch=1}}\text{.}
\end{equation}
With $l_{rate}$ normalized in $[0,1]$, it can work well together with $err_{val}$.

The last item in Eq.~(\ref{equ:score}), $sim$, represents the degree of similarity between a given architecture and each architecture we used for pretraining.
Moreover, we discover that given the same number of different \textit{module}s between two architectures, the place where the different \textit{module} lies is also one of the important factors for performance evaluation.
Besides, we also find that if we break up the continuity in relatively shallow places of the whole architecture, $acc_{val}$ may decrease slightly, but not that much if in relatively deep places.
Therefore, we define $sim$ through Eq.~(\ref{equ:sim1})-(\ref{equ:sim2}).
\begin{equation}
   sim=\frac{1}{c}f_{sim}(code,1,c)\label{equ:sim1}
   \setlength{\belowdisplayskip}{0.05cm}
\end{equation}
where
\begin{align}
   &f_{sim}(a,x,c) = \nonumber\\
   &\begin{cases}
      f_{sim}(a,x+1,c) + 1&x \le {c}\text{ and }a[x] = a[1]\\
      0&x > {c}\;\text{ or }\;a[x] \neq a[1]\\      
   \end{cases}\label{equ:sim2}
\end{align}
$code$ is the architecture encoding $i_{1}i_{2}...i_{c}$, for $c$ architectures in our knowledge base.

Our experiments show that the scoring function (Eq.~(\ref{equ:score})) has enough correlative relationship to help us accurately evaluate the performance for a given architecture when searching.

\section{Experiments}\label{Experiments}
Our experiments for the proposed ModuleNet are conducted with two stages, searching stage and evaluation stage.
The first is the searching stage.
As defined by Algorithm~\ref{alg:Overview}, we use $c=5, gen=30$ and $p_{size}=40$ for each experiment.
As for $n$ (Algorithm~\ref{alg:Overview}), $\alpha$ and $\beta$ (Eq.~(\ref{equ:score})), we will illustrate in the section of each experiment.
After searching stage, we can get the population of the final generation $pop_{final}$ (Algorithm~\ref{alg:Overview}). 
Then, in the second stage, we make all parameters in the architectures trainable and fine-tune the parameters.
Through this evaluation stage, we can determine the best final architecture for a given task.

As for classifier (depicted in Fig.~\ref{fig:uni-view}) in each experiment, we use three fully connected layers, with feature size of $input\text{ }size-4096-4096-class\text{ }number$, of which $input\text{ }size$ is determined by the last convolutional layer and $class\text{ }number$ is determined by the dataset.
We use the standard \textit{Cross Entropy Loss}, as
$$
\operatorname{loss}(x, class)=-\log\Big(\frac{\exp{x[class]}}{\sum_{j}\exp{x[j]}}\Big)\text{,}
$$
where $x$ is the array of network output, indicating possibilities to each class, and $class$ is the class label for input. 
We train for 20 epochs to make parameters in classifier convergent in the searching stage, and 50 epochs for fine-tuning in the evaluation stage.





\subsection{Result on CIFAR100}\label{sec:CIFAR100}
CIFAR100~\cite{cifar} is a highly used dataset for image classification.
Because it has a small image size, a deep model can be trained on this dataset in a short time.
Since its class number is more than CIFAR10~\cite{cifar}, it is broadly used in NAS scenario.

In this section, we use a configuration with $n$$=$$7$ architectures designed by craft for searching, containing ResNet34, ResNet50, ResNet101~\cite{ResNet} and VGG13, VGG16, VGG13bn, VGG16bn~\cite{VGG}.
We use the implements from \textit{torchvision}\footnote[1]{\ifx\unnamed\undefine Torchvision can be found in \url{https://github.com/pytorch/vision}, we \else We \fi are using version v0.3.0.} with their pretrained weights accordingly.
To make them compatible with the smaller input size, we remove the first pooling layer in ResNets.
Through parameter-free connections described in Sec.~\ref{Module Connection}, ModuleNet is able to search directly in macro space on this dataset.
For searching on CIFAR100 dataset, we use $\alpha=\beta=25$ to balance the performance evaluation.

\begin{figure}[htb]
   \centerline{
      \includegraphics[width=2.0in]{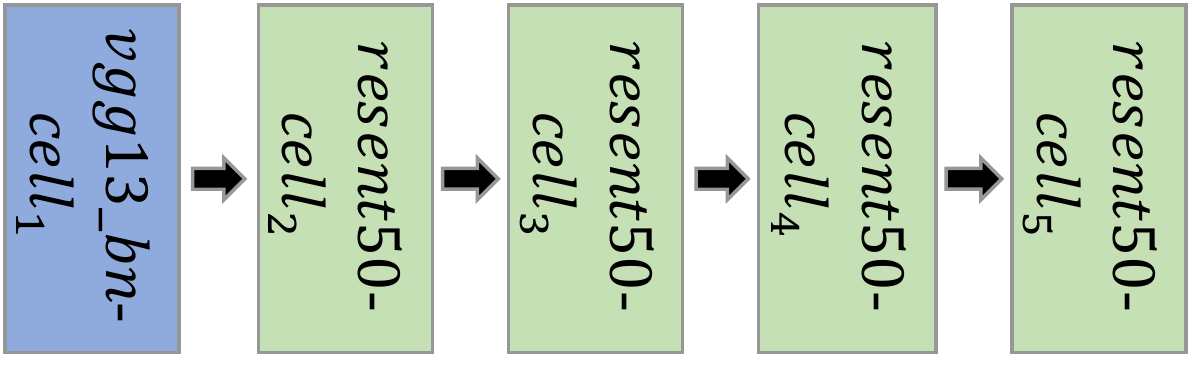}
   }
   \caption{Searched architecture based on craft design for CIFAR100: this architecture is searched on CIFAR100 with the knowledge base containing only VGGs and ResNets.}
   \label{fig:cifar100-craft}
   \vspace{-0.1cm}
\end{figure}

\begin{table}
   \small
   \begin{center}
   \begin{tabular}{|p{4cm}|c|}
   \hline
   Method & Test Error \\
   \hline\hline
   Best for ResNets + cutout & 22.97 \\
   Best for VGGs + cutout & 28.27 \\
   \hline
   ModuleNet (Ours) + cutout & \textbf{15.87} \\
   \hline
   \end{tabular}
   \end{center}
   \caption{Results comparison between our searched architecture and its origin architectures in the knowledge base on CIFAR100. In the fine-tune stage, we use $cutout\text{ }length=16$. Our dataset splitting for train/evaluation follows 40K/10K on training set for each architecture.}
   \label{tab:CIFAR100}
\end{table}

After two stages of searching, the best architecture we get is shown in Fig.~\ref{fig:cifar100-craft}.
Evaluation result comparison can be seen in Table~\ref{tab:CIFAR100}.
From the results, we may notice that although our searched architecture contains only one \textit{module} different from the original architecture, it can bring a huge increase in performance.

\subsection{Result on CIFAR10}
CIFAR10~\cite{cifar} is also a highly used dataset for image classification.
With fewer classes than CIFAR100, we could reduce the parameters in the classifier and fast evaluate the efficiency of our algorithm.
Taking advantage of short search time on CIFAR10, we go one step further in this section to prove the efficiency of our ModuleNet.
In Sec.~\ref{sec:cifar10-1}, we conduct the same experiment as on CIFAR100 and show our results.
In Sec.\ref{sec:cifar10-2}, we add some \textit{module}s whose architecture is searched by some state-of-the-art search algorithm to our knowledge base, and show that our algorithm can still make steady improvements even for these already-perform-well modules.

\subsubsection{Search with Craft Design}\label{sec:cifar10-1}
First, as a standard experiment similar to those on CIFAR100, we first fine-tune those architectures given in Sec.\ref{sec:CIFAR100} for new \textit{module}s in the knowledge base.
Then we apply searching with the same configurations as above.

\begin{figure}[htb]
   \centerline{
      \includegraphics[width=2.0in]{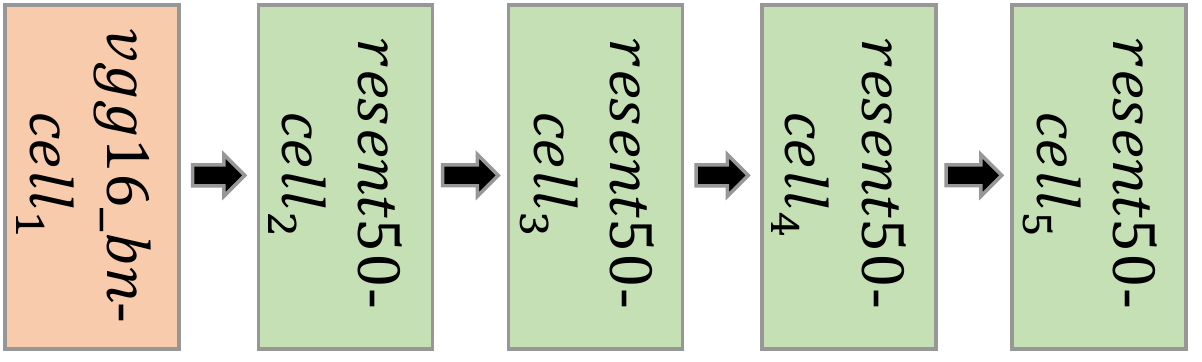}
   }
   \caption{Searched architecture based on craft design for CIFAR10: this architecture is searched on CIFAR10 with the knowledge base containing only VGGs and ResNets.}
   \label{fig:cifar10-craft}
   \vspace{-0.1cm}
\end{figure}

\begin{table}
   \small
   \begin{center}
   \begin{threeparttable}
      \begin{tabular}{|p{4cm}|c|}
      \hline
      Method & Test Error \\
      \hline\hline
      Best for ResNets\tnote{$\dagger$} & 6.43 \\
      Best for VGGs & 6.27 \\
      \hline
      ModuleNet (Ours) & \textbf{5.81} \\
      ModuleNet (Ours) + cutout\tnote{$\ddagger$} & \textbf{4.64} \\
      \hline
      \end{tabular}
      \begin{tablenotes}
         \footnotesize
         \item[$\dagger$] Obtained directly from \cite{ResNet}.
         \item[$\ddagger$] Data augmentation with $cutout\text{ }length=16$.
      \end{tablenotes}
   \end{threeparttable}
   \end{center}
   \caption{Results comparison between our searched architecture and its origin architectures in the knowledge base on CIFAR10. Our dataset splitting for train/evaluation follows 40K/10K on training set for each architecture.}
   \label{tab:CIFAR10}
\end{table}

After two stages of searching, the best architecture we get is shown in Fig.~\ref{fig:cifar10-craft} and results comparison can be seen in Table~\ref{tab:CIFAR10}.
From the results, we may have a similar conclusion as on CIFAR100, that a small change in \textit{module}s can lead to a slight improvement.
However, we also notice that different datasets need different architecture to guarantee better performance, and simply stacking the same cells or transferring architecture designed for other datasets may not bring the best results.

\subsubsection{Search with NAS Design}\label{sec:cifar10-2}
To extend our experiments, we then add some \textit{module}s searched by other NAS algorithms in our knowledge base to enlarge our search space.
DARTS~\cite{DARTS} is the first to introduce continuous relaxation to architecture representation, and famous work in NAS area.
By extending DARTS, PDARTS~\cite{PDARTS} and PC-DARTS~\cite{PC-DARTS} also make progress in searching for better architectures and follow the same configuration.
Therefore, we introduce \textit{module}s searched by these three algorithms \ifx\unnamed\undefine\footnote[1]{Implements can be found in \url{https://github.com/flymin/darts}} \fi into our knowledge base, referred to as DARTS, PDARTS, PC-DARTS-cifar (search by PC-DARTS for cifar) and PC-DARTS-image (search by PC-DARTS for ImageNet). Since we are using $c=5$ in other experiments, we also change their stacking strategy with reduction in 5, 10 and 15 layers, and 20 layers in total. In this way, we have 5 cells in each \textit{module}, togather with a stem cell as the first \textit{module}, making up 5 \textit{module}s.

\begin{figure}[htb]
   \centerline{
      \includegraphics[width=2.0in]{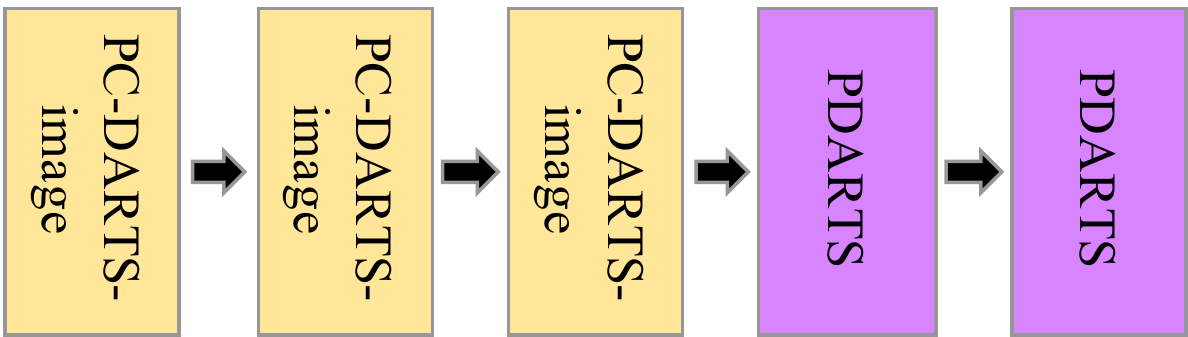}
   }
   
   \caption{Searched architecture based on other NAS results for CIFAR10.}
   \label{fig:cifar10-allNAS}
   \vspace{-0.1cm}
\end{figure}

Keeping other setting the same as previous experiments except for $\alpha=10,\beta=30$, after two stages of search, the architecture we get is shown in Fig.~\ref{fig:cifar10-allNAS} and results comparison can be seen in Table~\ref{tab:CIFAR10-2}.
From the results, we may notice that even for NAS searched architectures, stacking cells is not the best way.
Besides, our result does not contain those cells that perform better when stacked, from which we can conclude that where is no strong correlation on performance between single cell and whole architecture.

\begin{table}
   \small
   \begin{center}
   \begin{tabular}{|p{4cm}|c|}
   \hline
   Method & Test Error \\
   \hline\hline
   DARTSv2 + cutout & 2.86 \\
   PDARTS + cutout & 2.91 \\
   PC-DARTS-cifar + cutout & 2.87 \\
   PC-DARTS-image + cutout & 2.8 \\
   \hline
   ModuleNet (Ours) + cutout  & \textbf{2.77}\\
   \hline
   \end{tabular}
   \end{center}
   \caption{Results comparison between our searched architecture and its origin architectures in the knowledge base on CIFAR10. Search space of this experiment contains \textit{module} searched by NAS algorithms. We use all 50K training images to train and testing split for validation, following the training scheme in DARTS~\cite{DARTS}.}
   \label{tab:CIFAR10-2}
\end{table}

\subsubsection{Experiments on ImageNet}
To evaluate the architecture searched by our approach on larger datasets, we ran the architecture directly on ImageNet using DGX station. From Fig. \ref{fig:cifar100-craft} and Fig. \ref{fig:cifar10-craft}, we can see that the best architectures searched by ModuleNet replace the first module in ResNet with VGG's. We suppose a possible explanation that in shallow layers, modules need to rule out more useless information, whereas in deep layers, with losing of useless information, modules need to be more careful when filtering. Therefore, VGG modules, which are better at ruling out information, are used as shallow layers. Whereas ResNet modules, which are better at identifying and keeping useful information, are used as deep layers. In order to evaluate the explanation, we conduct experiments on ImageNet: the first module of VGG13bn replacing ResNet-50 (VGG13bn+ResNet-50) and ResNet-101 (VGG13bn+ResNet-101), and results are shown in Table \ref{tab:imagenet} . 

We inherit the knowledge of VGG13bn, Resnet-50 and Resnet-101 models which are trained in the ImageNet dataset \cite{resnetResult}. The searched architectures transformed into ImageNet also outperforms VGG13bn, Resnet-50 and Resnet-101 in top1 and top5 test error. These results show the searched architecture replacing the first module of ResNet with VGG's is also robust for the large-scale dataset ImageNet.

\begin{table}
   \small
   \begin{center}
   \begin{tabular}{|p{3.1cm}|c|c|}
   \hline
   Method & Test Error top1 &Test Error top5\\
   \hline
   \hline
VGG13bn & 28.45 &9.63 \\
\hline
   \hline
    ResNet-50 & 23.85 &7.13 \\
   \hline
   VGG13bn+ResNet-50 & \textbf{22.744} &\textbf{6.396}\\
   \hline
   \hline
ResNet-101 &22.63 &6.44\\
   \hline
   VGG13bn+ResNet-101 & \textbf{21.308}&\textbf{5.802}\\
   \hline
   \end{tabular}
   \end{center}
   \caption{Results comparison between our searched architecture and its origin architectures in the knowledge base on ImageNet. }
   \label{tab:imagenet}
\end{table}

\section{Ablation Studies}
In this section, we will show some additional experiments to prove the effectiveness of three core parts in our proposed method separately. 
\subsection{Efficiency of Performance Evaluation}
As a core part of our search algorithm, $score$ in Eq.~(\ref{equ:score}) performs an important role to make a comparison between different architectures during evolution.
Therefore, the efficiency of the evaluation function is very important and directly related to the final results we get.
In this section, we conduct some more experiments to prove the efficiency of our evaluation function $score$.

As a NAS algorithm aiming to improve the performance on image classification tasks, the basic evaluation metric should be \textit{Test Error}.
However, directly calculating Test Error requires fully trained of architecture on the training set, which demands large computing cost and time.
Therefore, we introduce a new strategy and new function to evaluate.
Accordingly, the best way to show the efficiency of our strategy and function is to make a comparison with Test Error for architectures after fully trained.

\begin{figure}[htb]
   \centerline{
      \includegraphics[width=\linewidth]{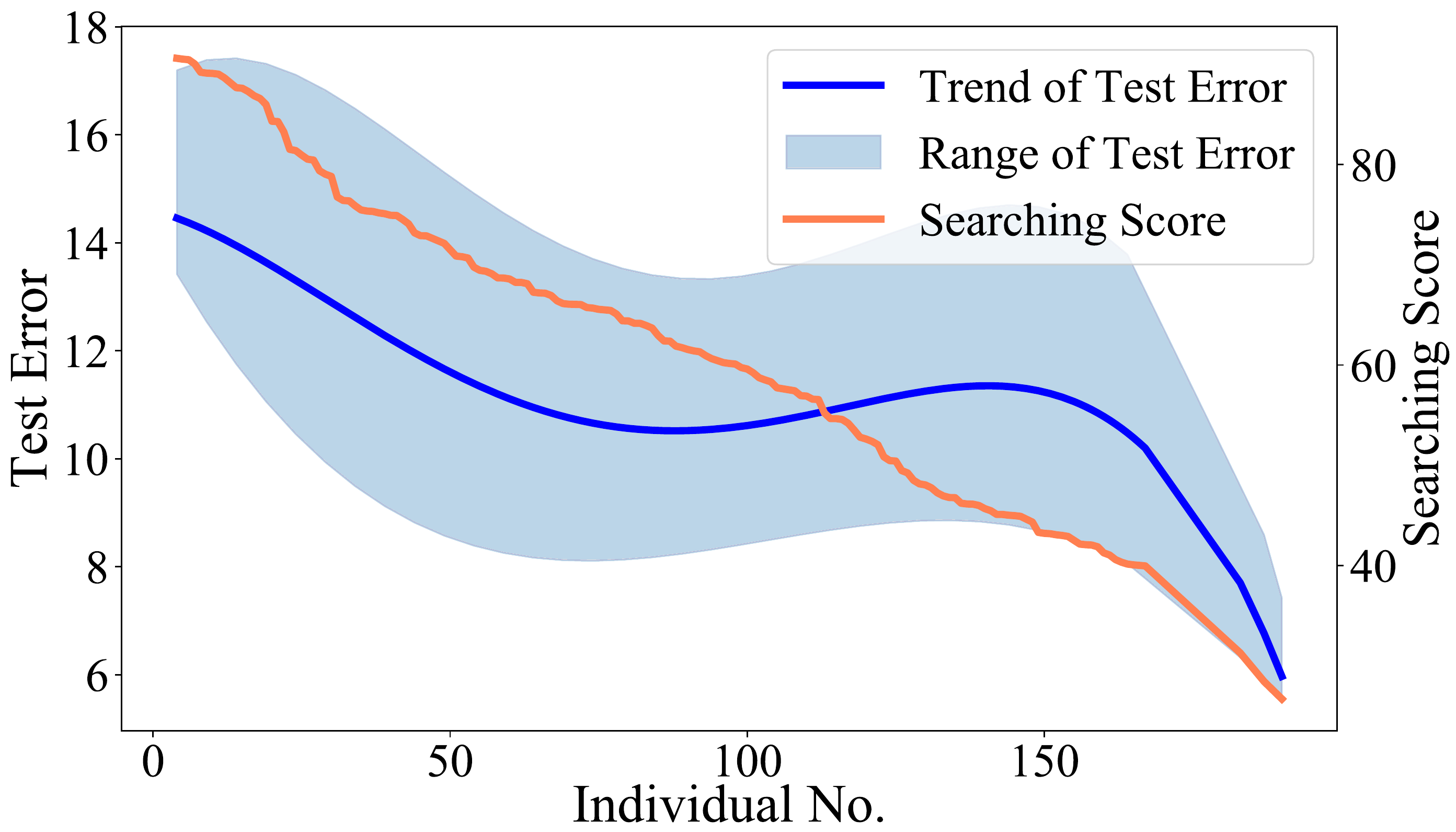}
   }
   \caption{Comparison between results generated through our searching strategy by $score$ (donated by \textit{Searching Score}), and \textit{Test Error} after retraining. Both metrics are the lower the better. Architectures are obtained from all 30 generations of populations in the experiment of Sec.\ref{sec:cifar10-1}, descendingly sorted according to \textit{Searching Score}.}
   \label{fig:cifar10_new_compare}
\end{figure}

As shown in Fig.~\ref{fig:cifar10_new_compare}, descending of \textit{searching score} indicates descending of \textit{Test Error}.
Such a result can prove that there is a correlative relationship between our evaluation strategy and test error, which proves that our strategy is usable to search for better architectures.
Furthermore, architectures we used in Fig.~\ref{fig:cifar10_new_compare} are from the same searching path of experiment in Sec.\ref{sec:cifar10-1}, which can be a side proof that our evaluation strategy is fit for the evolution algorithm we used.
Even though, we make a direct proof for the efficiency of evolution algorithm in the following section.

\subsection{Efficiency of Evolution Search}
In general, the evolution algorithm NSGA-II, is a multi-objective optimization algorithm.
We choose NSGA-II as back end to bring more scalability of our search algorithm.
In future, other objectives, such as the amount of parameters, latency, or amount of floating-point operations, can be easily extended into current search framework.
However, for now, we only consider $score$ as the only objective.

\begin{figure}[htb]
   \vspace{-0.2cm}
   \centerline{
      \includegraphics[width=0.9\linewidth]{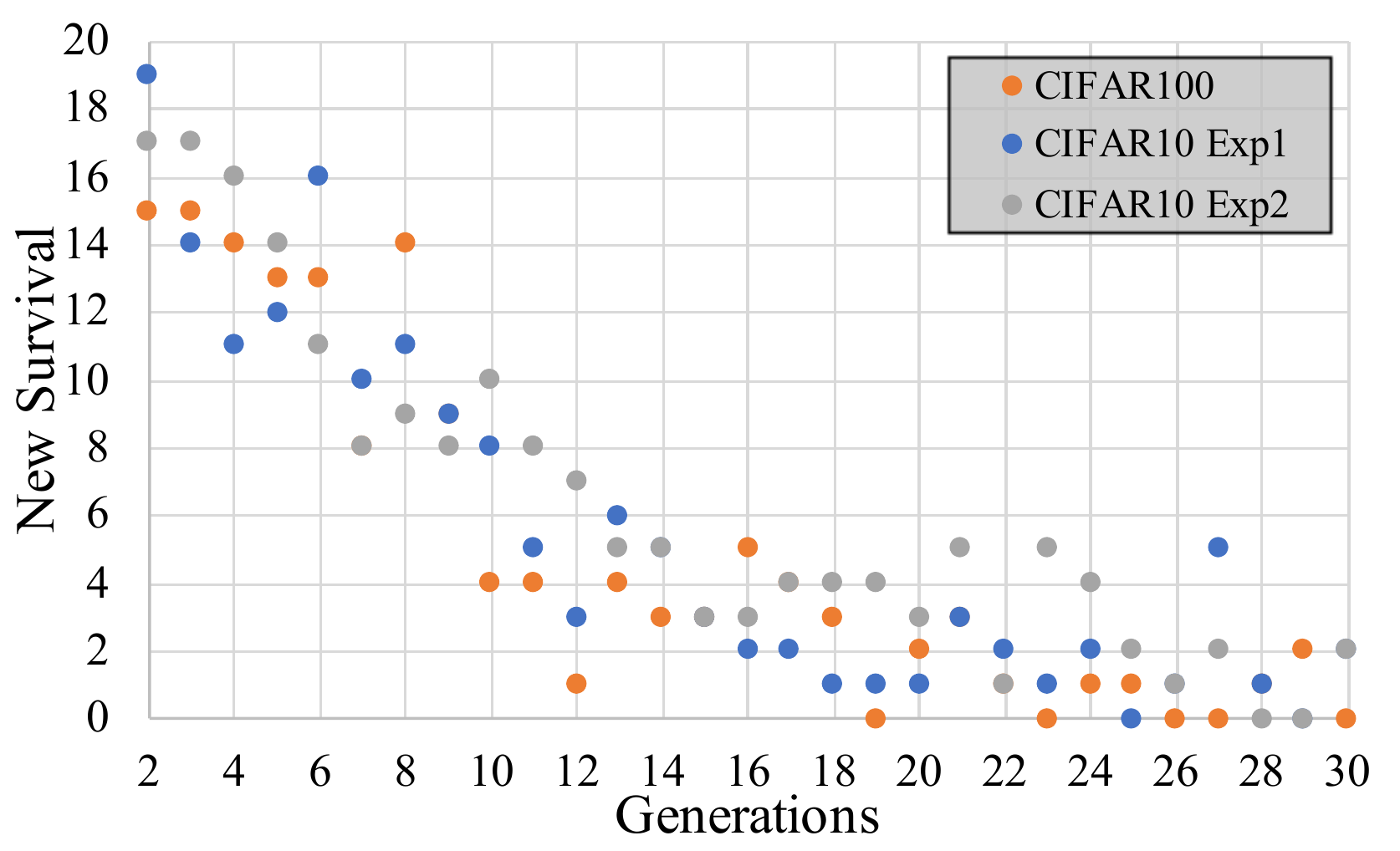}
   }
   \caption{New survival between generations: For each experiment in Sec.\ref{Experiments}, new survived genotypes for $pop_{i}$ (population of generation $i$) comparing to $pop_{i-1}$.}
   \label{fig:survival}
   \vspace{-0.1cm}
\end{figure}

To evaluate the efficiency of evolution algorithm, one side is judging final result searched by this algorithm, which has already shown in Sec.\ref{Experiments}; the other side is convergence of the algorithm.
Although \textit{mutation} and \textit{crossover} is made between generations, we still expected that genotypes of living generation may become relatively stable after generations of evolution.
Therefore, we calculate genotype changes between (indicated by \textit{new survival}) two consecutive generations, as shown in Fig.~\ref{fig:survival}.

We can notice that in each separate experiment, genotypes in populations will always converge to be stable after generations of evolution.
From such results, we can conclude that the evolution algorithm, NSGA-II, is fit for NAS task within the scenario of our proposed ModuleNet.

\subsection{Non-parameter Connection}
As illustrated in Sec.~\ref{Module Connection}, using non-parameter connections can slightly reduce trainable parameters when searching, and thus reduce the searching time.
Although we theoretically keep enough diversity when connecting the preceding \textit{module} with the following \textit{module}, these operations, however, may leave a question.
Do non-parameter connections result in a negative effect on the performance judgment?
To prove the efficiency of our proposed non-parameter connections, we do some further experiments.

\begin{figure}[htb]
   \centerline{
      \includegraphics[width=3.2in]{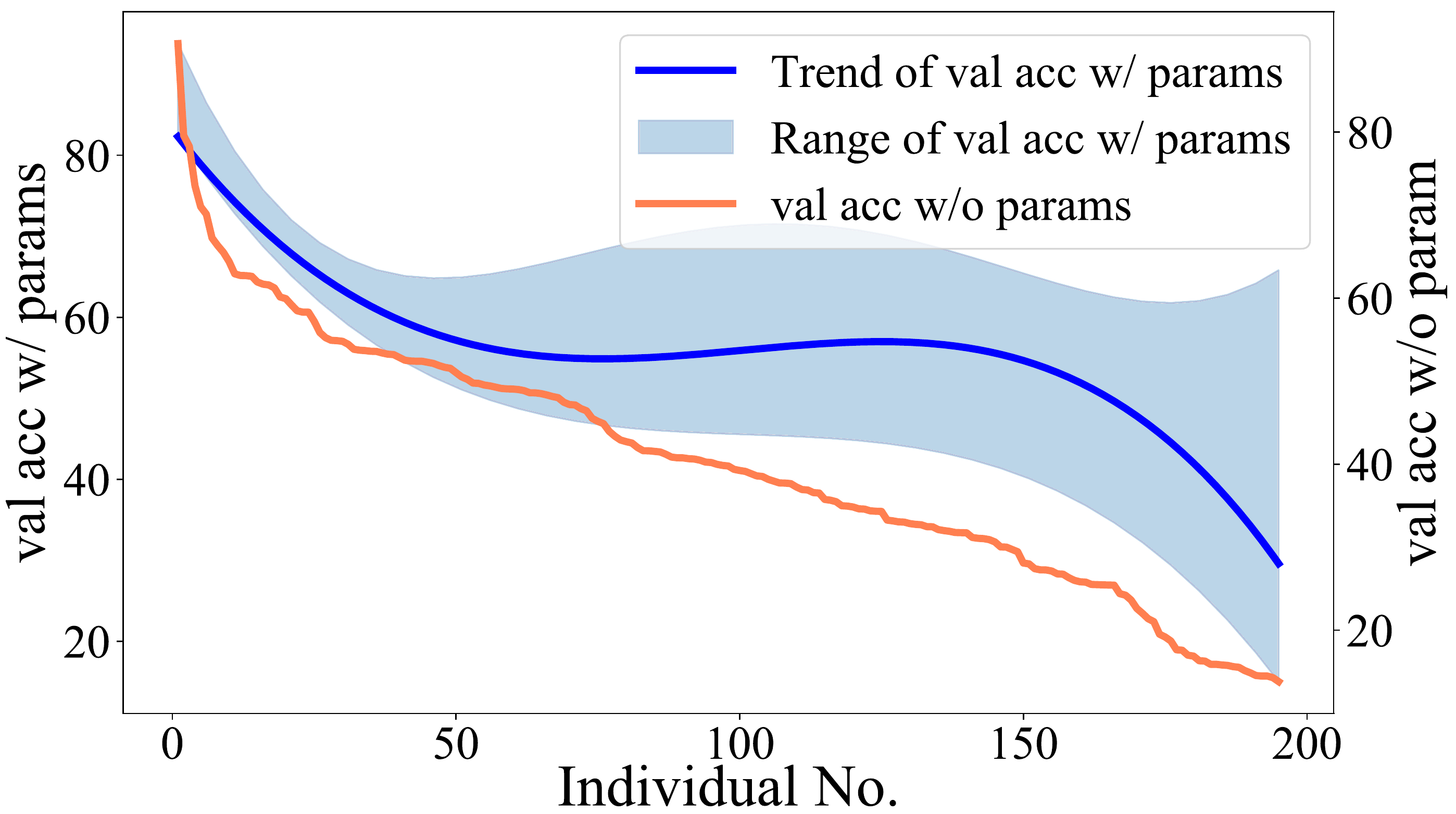}
   }
   \caption{Comparison between validation accuracy (\textit{val acc}) generated through non-parameter connections (\textit{val acc w/o params}, described in Sec~\ref{Module Connection}) and connections using $1$$\times$$1$ convolutions (\textit{val acc w/ params}). Architectures are obtained from all 30 generations of populations in experiment of Sec.\ref{sec:cifar10-1}, descendingly sorted by \textit{val acc w/o params}.}
   \label{fig:cifar10_new_compare_con}
   \vspace{-0.2cm}
\end{figure}

In our experiments, we apply a $1$$\times$$1$ convolution between each \textit{module} to transfer between different channels.
$1$$\times$$1$ convolution can be the simplest way to change channel dimensions and keep others.
Besides, it contains trainable parameters to make it adjustable by gradient.

We use architectures searched from all 30 generations of populations in experiment of Sec.\ref{sec:cifar10-1}, and make comparison with performance under search setting (fix parameters in each \textit{module} and tune others).
As shown in Fig.~\ref{fig:cifar10_new_compare_con}, using non-parameter connections can keep, and even strengthen the differential ability of our algorithm.
Specifically, for those better architecture (left end in X axis), both metrics indicate better results, and vice versa.
Besides, for those architectures in the middle of X axis, non-parameter connections could lead to a better differential status.

Therefore, our non-parameter connection is not only usable, but better fit for our search algorithm.

\section{Conclusions}
This paper presents ModuleNet, a new NAS algorithm to fully inherit existing knowledge and explore for new architecture design.
We propose that both architecture and trained parameters of an existing model should be used for further exploration.
By decomposing existing architectures into \textit{module}s, we can use a uniform-view to reorganize and rediscover on them.
In this way, we can make CNNs transferred quickly among different tasks and datasets, and always guarantee a performance improvement.

In our experiments, we not only show that the search architecture has better performances, but show the efficiency of our \textit{score} equation, evolution algorithm and connections between \textit{module}s.
All of these prove that existing knowledge is of great importance, and ModuleNet has set up a new NAS scheme for using them.
Actually, there are also many directions to improve ModuleNet further.
For example, \textit{score} equation can only indicate a relevance relationship, which can be better if a linear relationship is reached.
And some extensions may be added to the search algorithm to fulfill other constrictions.
These interesting topics could be potential directions for future studies.








\bibliographystyle{IEEEtran}
\bibliography{refs}

\end{document}